\documentclass[conference]{IEEEtran}
\IEEEoverridecommandlockouts

\usepackage[T1]{fontenc}
\usepackage{graphicx}
\usepackage{amsmath}
\usepackage{amssymb}
\usepackage{amsfonts}
\usepackage{booktabs}
\usepackage{float}
\usepackage{stfloats}  
\usepackage{enumitem}
\usepackage{caption}
\usepackage{url}
\usepackage{cite}
\usepackage{xcolor}
\usepackage{hyperref}

\hypersetup{
  hypertexnames=false,
  colorlinks=true,
  linkcolor=blue!50!black,
  citecolor=blue!50!black,
  urlcolor=blue!50!black
}

\usepackage{fancyhdr}
\newcommand{\footerlabel}{Preprint --- arXiv v2 (November 2026)}
\fancypagestyle{plain}{\fancyhf{}\fancyfoot[C]{\footnotesize \footerlabel \hfill \thepage}}
\setlist[itemize]{leftmargin=*}

\begin{document}

\title{What Does a Cosine of 0.8 Mean?\\
Monotone Calibration of Embedding Similarity}

\author{\IEEEauthorblockN{Nicolas Tacheny}
\IEEEauthorblockA{University of Mons (UMONS), Belgium \\
nicolas.tacheny@umons.ac.be}}

\maketitle
\thispagestyle{plain}  

\begin{abstract}
Text embedding models turn sentences into vectors, and the cosine between two vectors is the usual measure of their similarity. Search, semantic caching and deduplication often accept a pair when this cosine reaches a fixed threshold. When people judge one pair of sentences more similar than another, the cosine usually agrees, but its value means different things for different models, because each model packs its scores into its own narrow range. The same threshold can therefore keep most pairs with one model and few with another.

We propose to calibrate the score instead of changing the model. An increasing function fitted on sentence pairs rated by people translates the cosine of each model into the human rating scale and never reverses the order of two pairs.

We fit three such functions for six models, ranging from a simple S-shaped curve to a free step function. On new pairs, calibrated scores agree with human ratings and the models agree with one another. One threshold selects pairs of similar quality whatever the model. The scale obtained belongs to the human ratings used to fit it, so its values shift on two datasets rated differently while the models stay comparable.
\end{abstract}


\section{Introduction}
\label{sec:intro}

Text embedding models represent a sentence as a vector, and the cosine between two vectors is the standard measure of how similar two sentences are. Many systems act on this value through a threshold. A semantic cache reuses a stored answer when a new question is close enough to an old one, and deduplication merges two texts when their cosine is high. Such a threshold assumes that a cosine of $0.8$ means the same thing whatever the model. It does not. Trained encoders crowd their vectors into a narrow region of the space, so their scores cluster in a high band whose position depends on the encoder. On the same sentence pairs, a cosine of $0.8$ corresponds to a human rating of $0.763$ with one encoder and $0.189$ with another. The order of pairs survives this, but the values cannot be read on their own or compared across encoders.

Recent work recognizes this limitation. A review of cosine similarity describes it as bounded but uncalibrated and notes that a threshold learned for one model need not transfer to another~\cite{you2026semantics}. In semantic caching, Schroeder et al.~\cite{schroeder2026vcache} show that no single fixed threshold guarantees that a reused answer is correct, and they learn a separate threshold for each cached entry.

Remedies proposed so far transform the embedding space itself, which forces every embedding to be recomputed. We propose to \emph{calibrate the metric, not the space}, as one calibrates a measuring instrument: its readings are compared with a reference and a correction is attached to it. For each encoder we fit an increasing map on sentence pairs rated by people. The map puts the cosine on the human scale and, being increasing, never reverses the order of two pairs. Our claim concerns calibration itself and not a particular map, so we check every result with three maps of increasing flexibility: a sigmoid, beta calibration and isotonic regression.

We test this on six common encoders, fitting the maps on the STS Benchmark and measuring on pairs not used for fitting. Three findings support our thesis that calibration makes cosine similarity comparable across encoders, on the scale of the human judgments used to fit it:
\begin{enumerate}
  \item \textbf{Same value, different meaning.} The encoders order pairs almost equally well, but their raw values depart from human ratings in different ways.
  \item \textbf{Most of the difference is a matter of scale.} After calibration every encoder agrees with human ratings, the scores of two encoders for the same pair come close, and one threshold selects pairs with nearly the same precision whatever the encoder. The small gap that remains comes from pairs that the encoders order differently.
  \item \textbf{The scale belongs to the reference judgments.} On two datasets rated under other instructions, the calibrated values no longer match human ratings, but the encoders remain comparable with one another.
\end{enumerate}

After related work in Section~\ref{sec:related}, Section~\ref{sec:setting} describes the encoders, the data and the metrics. Section~\ref{sec:meaning} compares raw cosine with human judgments, Section~\ref{sec:calibration} introduces the calibration and states how we test it. Sections~\ref{sec:agreement} to~\ref{sec:protocol} show what it achieves and where it stops. Applications and limits close the paper in Section~\ref{sec:discussion}.

\section{Related Work}
\label{sec:related}

Work on the problem began with the geometry of embeddings. Ethayarajh~\cite{ethayarajh2019contextual} showed that representations from BERT, ELMo and GPT-2 grow increasingly anisotropic in upper layers so that even unrelated items get high cosine similarity. Gao et al.~\cite{gao2019representation} found the word embeddings of language generation models crowded into a narrow cone and called this the representation degeneration problem. Wang and Isola~\cite{wang2020understanding} proved that the contrastive loss asymptotically optimizes the alignment of positive pairs and the uniformity of the whole distribution on the hypersphere. Whitening~\cite{su2021whitening} and contrastive training spread the representations more evenly by modifying them, so every stored embedding must be recomputed.

Two recent papers look instead at the thresholds applied to the cosine. A review of cosine similarity~\cite{you2026semantics} describes it as uncalibrated, notes that a threshold learned for one model need not transfer to another and advises calibrating separately when thresholds matter. Schroeder et al.~\cite{schroeder2026vcache} observe that semantic caches use a predefined threshold such as $0.8$ that gives no formal guarantee of correctness, and they learn a separate threshold for each cached prompt.

Calibration itself comes from classification, where Platt~\cite{platt2000probabilistic} introduced sigmoid calibration for support vector machines. Zadrozny and Elkan~\cite{zadrozny2002transforming} used isotonic regression to calibrate classifier probabilities and Niculescu-Mizil and Caruana~\cite{niculescumizil2005predicting} compared both across classifiers. Kull et al.~\cite{kull2017beta} proposed beta calibration, a family with three parameters that avoids the mismatch of the sigmoid when scores are not normally distributed. Guo et al.~\cite{guo2017calibration} showed that modern neural networks are poorly calibrated despite high accuracy. Desai and Durrett~\cite{desai2020calibration} found pretrained transformers calibrated in domain but less so out of domain. All of this work targets classification confidence.

We follow the advice of the review and carry calibration over to similarity, with the graded human judgments of the STS Benchmark~\cite{cer2017semeval} as the target, a dataset that MTEB~\cite{muennighoff2023mteb} uses to evaluate embeddings. Unlike whitening it leaves the space intact, and unlike a threshold per cached entry it gives every encoder one scale on which scores can be compared.

\section{Encoders, Data and Metrics}
\label{sec:setting}

An encoder $\phi$ maps a text $a$ to a vector $\phi(a) \in \mathbb{R}^d$, which we normalize to $\psi(a) = \phi(a)/\|\phi(a)\|_2$ on the unit sphere $\mathbb{S}^{d-1}$. The raw similarity of two texts is the inner product of their normalized embeddings, which equals the cosine of their angle:
\begin{equation}
s(a, b) = \langle \psi(a), \psi(b) \rangle \in [-1, 1].
\end{equation}
If embeddings spread uniformly over the sphere, the cosine of two unrelated texts would concentrate around $0$, with deviations of order $1/\sqrt{d}$~\cite{vershynin2018high}. Trained encoders do not spread their embeddings in this way. They crowd them into a narrow cone, as Ethayarajh~\cite{ethayarajh2019contextual} found for BERT, ELMo and GPT-2, so that unrelated texts still receive a large positive cosine whose level depends on the encoder. Training choices set this level too. E5, for instance, is trained with a contrastive temperature of $0.01$~\cite{wang2022text}, and $82.6\%$ of its STS-test cosines reach $0.8$ (Table~\ref{tab:raw}).

We study six encoders, chosen for their wide use and their different training. \texttt{paraphrase-mpnet-base-v2} is trained on paraphrase pairs, \texttt{all-MiniLM-L6-v2} ($d = 384$) and \texttt{all-mpnet-base-v2} on large collections of sentence pairs. The three others are recent general-purpose encoders, \texttt{bge-base-en-v1.5} (CLS pooling), \texttt{e5-base-v2} (with its \texttt{query:} input prefix) and \texttt{gte-base}. All except \texttt{all-MiniLM-L6-v2} have $d = 768$. They run locally in ONNX format through Transformers.js~2.17.2\footnote{\url{https://github.com/huggingface/transformers.js}} with the pooling each was trained with, which keeps every experiment reproducible without external APIs. A text of several sentences is embedded sentence by sentence. The sentence embeddings are then averaged with softmax weights $w_k = \exp(c_k) / \sum_j \exp(c_j)$ given by the cosine $c_k$ of sentence $k$ to the mean embedding, and the result is normalized. This concerns $15$ of the $2{,}758$ texts in STS-test, $998$ of the $10{,}462$ in STR-2022 and none in SICK-R.

Human judgments come from the STS Benchmark, which annotates sentence pairs with similarity scores on a $0$--$5$ scale that we rescale to $s^{(h)} \in [0,1]$. We call its train split of $5{,}749$ pairs STS-train and its test split STS-test.

Two numbers summarize agreement with human judgments. The first is read on a reliability diagram, which groups pairs by score and plots, for each group, the mean human score against the mean model score. A calibrated encoder lies on the diagonal. The expected calibration error (ECE) is the average distance between the curve and the diagonal, weighted by the number of pairs in each of $B = 10$ equal-width bins $B_b$:
\begin{equation}
\label{eq:ece}
\mathrm{ECE} = \sum_{b=1}^{B} \frac{|B_b|}{N} \Big| \overline{s^{(h)}}_{B_b} - \overline{s}_{B_b} \Big|,
\end{equation}
where $N$ is the number of pairs, $\overline{s}_{B_b}$ the mean model score in bin $B_b$ and $\overline{s^{(h)}}_{B_b}$ the mean human score. A second number, the Spearman correlation $\rho$ between model and human scores, completes the picture. It measures agreement in order, which no strictly increasing transformation of the scores can change.

\section{Same Value, Different Meaning}
\label{sec:meaning}
We first compare raw cosine with human judgments, before any calibration.

\begin{figure}[H]
\centering
\includegraphics[width=\columnwidth]{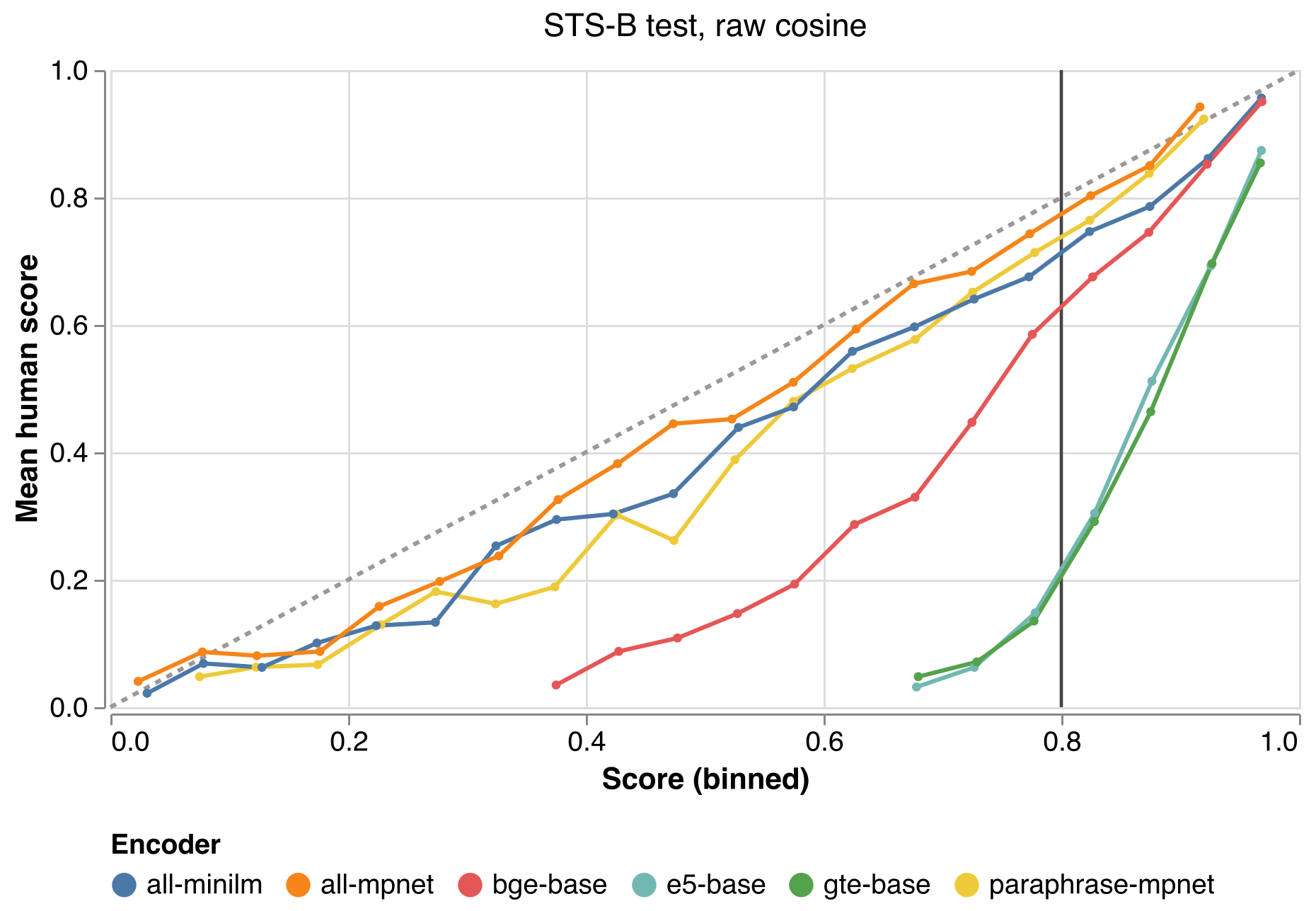}
\caption{Reliability diagram of the six encoders, raw cosine on STS-test. Each point is the mean human score of the pairs in a score bin of width $0.05$ (bins with at least $15$ pairs). The dotted diagonal marks perfect calibration and the vertical line a score of $0.8$.}
\label{fig:reliability-raw}
\end{figure}

The six encoders order pairs almost equally well, with Spearman correlations between $0.813$ and $0.864$ on STS-test (Table~\ref{tab:raw}). Their values, however, depart from the human scale. In the reliability diagram of Fig.~\ref{fig:reliability-raw}, every curve lies below the diagonal, which means that every encoder overestimates similarity. It does so slightly for \texttt{all-mpnet-base-v2} and strongly for \texttt{e5-base} and \texttt{gte-base}, whose ECE reaches $0.359$ (Table~\ref{tab:raw}).

\begin{table}[H]
\centering
\footnotesize
\setlength{\tabcolsep}{4pt}
\begin{tabular}{lccc}
\toprule
\textbf{Encoder} & $\boldsymbol{\rho}$ & \textbf{ECE} & \textbf{Cosine $\geq 0.8$} \\
\midrule
paraphrase-mpnet & 0.861 & 0.087 & 26.1\% \\
all-MiniLM       & 0.819 & 0.084 & 26.9\% \\
all-mpnet        & 0.813 & 0.042 & 17.5\% \\
bge-base         & 0.859 & 0.220 & 41.8\% \\
e5-base          & 0.864 & 0.354 & 82.6\% \\
gte-base         & 0.854 & 0.359 & 84.5\% \\
\bottomrule
\end{tabular}
\caption{Raw cosine of six encoders on STS-test. $\boldsymbol{\rho}$: Spearman correlation with human scores. \textbf{Cosine $\geq 0.8$}: share of the $1{,}379$ test pairs whose cosine reaches $0.8$. Humans rate $24.5\%$ of these pairs at least $0.8$.}
\label{tab:raw}
\end{table}

Fig.~\ref{fig:reliability-raw} also shows what a raw cosine means for each encoder. Test pairs with a cosine near $0.8$ are rated on average $0.699$ to $0.763$ by humans for \texttt{paraphrase-mpnet-base-v2}, \texttt{all-MiniLM-L6-v2} and \texttt{all-mpnet-base-v2}, $0.633$ for \texttt{bge-base} and only $0.189$ to $0.209$ for \texttt{e5-base} and \texttt{gte-base}. The same raw value thus means very similar for some encoders and barely related for others.

How much an encoder overestimates depends on where it places its scores. Humans rate $24.5\%$ of the test pairs at least $0.8$. Yet a threshold of $0.8$ on raw cosine accepts $17.5\%$ of the pairs with \texttt{all-mpnet-base-v2}, and $82.6\%$ and $84.5\%$ with \texttt{e5-base} and \texttt{gte-base} (Table~\ref{tab:raw}). A threshold of $0.8$ is thus demanding for \texttt{all-mpnet-base-v2} and lax for \texttt{e5-base} and \texttt{gte-base}. Such a raw threshold therefore cannot be read in absolute terms, nor carried over from one encoder to another.

\section{Calibrating the Metric}
\label{sec:calibration}

Section~\ref{sec:meaning} shows that the encoders get the order of pairs right and their values wrong. What needs correcting is therefore how a score reads, while the geometry that produces the order can stay as it is. The remedies discussed in Section~\ref{sec:related} act on the embedding space, by whitening it or by fine-tuning the encoder, after which every embedding must be recomputed and every index rebuilt. We act on the score and \emph{calibrate the metric, not the space}. The embeddings and their cosine stay exactly as they are. A map applied to the cosine changes only the number reported for each pair.

The reference is human judgment. For each encoder $e$ we learn its own map $g_e$ from raw cosine to the human scale and replace $s$ by the calibrated similarity $\tilde{s} = g_e(s)$. Since each map concerns a single encoder, we drop the index $e$ in the rest of this section. The natural target for $g(s)$ is the mean human score of the pairs that receive cosine $s$, that is $\mathbb{E}[s^{(h)} \mid s]$. We estimate it by fitting $g$ on $n$ rated pairs so as to minimize the squared error
\begin{equation}
\sum_{i=1}^{n} \big( s^{(h)}_i - g(s_i) \big)^2,
\end{equation}
where $s_i$ is the cosine of pair $i$ and $s^{(h)}_i$ its human score. A calibrated score thus reads directly on the human scale, close on average to the score people give. The map is fitted once on rated pairs and then applies to any score the encoder produces.

We only consider increasing maps because they keep the order of pairs and the encoders already get this order right. If $g$ is strictly increasing, then for any texts $x, y, z$,
\[
s(x,y) \geq s(x,z) \iff \tilde{s}(x,y) \geq \tilde{s}(x,z).
\]
A search for the nearest neighbors of a text therefore returns the same texts in the same order after calibration. Filtering pairs with a threshold also keeps the same pairs, once the raw threshold $\tau$ is replaced by its calibrated value $g(\tau)$. A non-decreasing map with flat steps may merge distinct scores into ties but never reverses their order. By contrast, a map that is not monotone, such as a high-degree polynomial, can reverse the order of two pairs, however well it fits.

The argument so far fixes the class of maps, not one member of it. Our claim concerns calibration itself and not one particular map. We therefore use three increasing maps that range from the most constrained to the most flexible. The first map is the sigmoid of Platt scaling~\cite{platt2000probabilistic},
\begin{equation}
\tilde{s} = g(s) = \frac{1}{1 + e^{-(a s + b)}}.
\end{equation}
Its two parameters match what Section~\ref{sec:meaning} shows the encoders differ in. The slope $a$ sets how much the band of an encoder is stretched and the offset $b$ sets where its center falls. This center is the cosine $s = -b/a$ at which $g(s) = 0.5$. The second map is beta calibration~\cite{kull2017beta}. It has three parameters that let the map rise at a different rate near the top of the scale than near the bottom. With $u = (s+1)/2$,
\begin{equation}
g(s) = \frac{1}{1 + e^{-(c + a \ln u - b \ln (1 - u))}}, \quad a, b > 0.
\end{equation}
Both maps are strictly increasing and stay within $(0, 1)$. The third map is isotonic regression~\cite{zadrozny2002transforming} and imposes no form at all. It fits the best non-decreasing step function and its steps merge distinct cosines into ties. Each map is fitted on STS-train, separately for each encoder. For the sigmoid and beta calibration, the parameters are found by the Levenberg--Marquardt algorithm~\cite{levenberg1944method,marquardt1963algorithm}, which adjusts them step by step until the squared error stops decreasing, and isotonic regression is found exactly by the pool-adjacent-violators algorithm~\cite{ayer1955empirical}.

Our claim is that calibration makes cosine scores comparable across encoders on the scale of the human judgments used to fit it. Every map is fitted on STS-train and each result is reported for the three maps on pairs not seen during fitting. We test the claim in three steps. Section~\ref{sec:agreement} tests whether calibrated scores agree with human judgments, Section~\ref{sec:scale} whether they make the encoders agree with one another and Section~\ref{sec:protocol} whether they carry over to judgments collected differently.

\begin{figure*}[b]
\centering
\includegraphics[width=\textwidth]{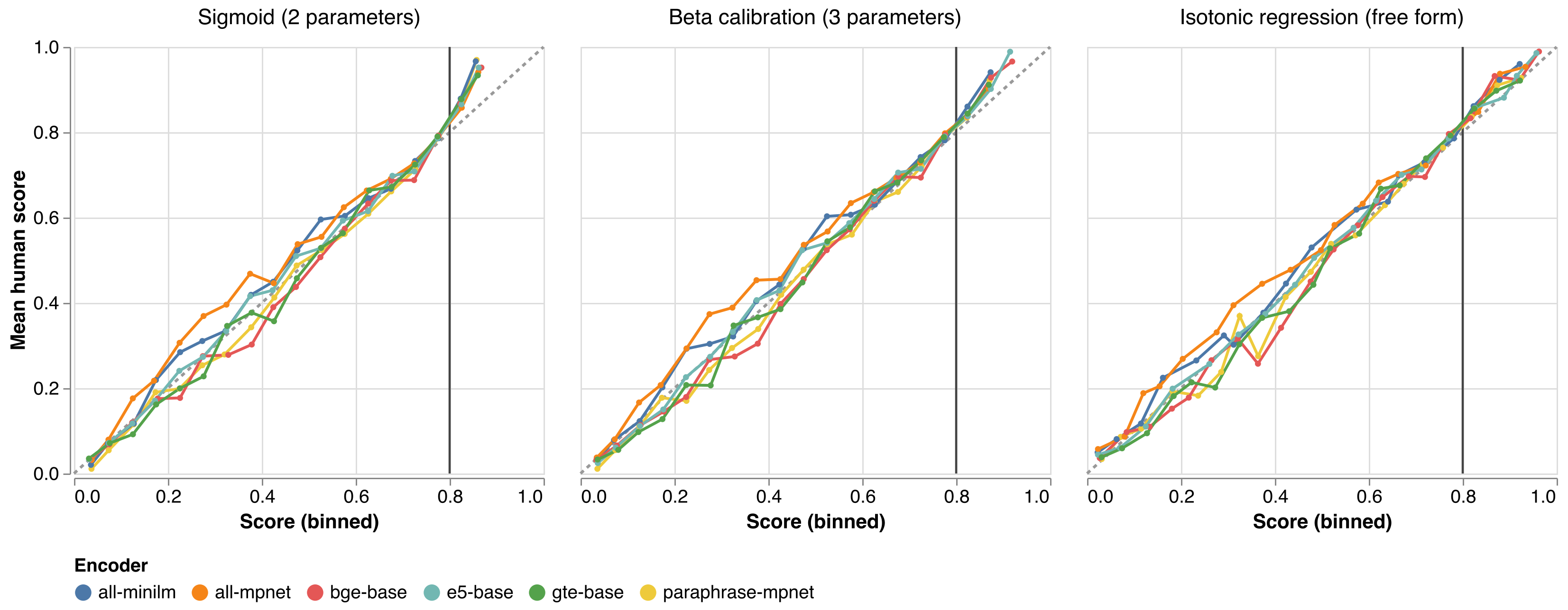}
\caption{Reliability diagram of the six encoders on STS-test after calibration by each of the three maps, fitted on STS-train. Each point is the mean human score of the pairs in a score bin of width $0.05$ (bins with at least $15$ pairs). The dotted diagonal marks perfect calibration and the vertical line a score of $0.8$.}
\label{fig:reliability-calibrated}
\end{figure*}

\section{Calibrated Scores Agree with Human Judgments}
\label{sec:agreement}

The first test asks whether calibrated scores agree with human judgments on new pairs from the same benchmark. For each encoder we read the reliability diagram of its calibrated scores on STS-test (Fig.~\ref{fig:reliability-calibrated}) and summarize it by the ECE (Table~\ref{tab:calibrated}). We also check that calibration keeps the order of pairs by the change in $\rho$.

\begin{table}[H]
\centering
\footnotesize
\setlength{\tabcolsep}{5pt}
\begin{tabular}{lccccc}
\toprule
 & & \multicolumn{4}{c}{\textbf{Calibrated ECE}} \\
\cmidrule(lr){3-6}
\textbf{Encoder} & $\boldsymbol{\Delta\rho}$ & \textbf{Sigmoid} & \textbf{Beta} & \textbf{Isotonic} & \textbf{Range} \\
\midrule
paraphrase-mpnet & $+0.0001$ & 0.018 & 0.014 & 0.014 & 0.003 \\
all-MiniLM       & $-0.0002$ & 0.025 & 0.024 & 0.026 & 0.002 \\
all-mpnet        & $-0.0003$ & 0.039 & 0.038 & 0.039 & 0.001 \\
bge-base         & $0.0000$ & 0.025 & 0.022 & 0.022 & 0.003 \\
e5-base          & $-0.0007$ & 0.016 & 0.017 & 0.013 & 0.004 \\
gte-base         & $0.0000$ & 0.021 & 0.022 & 0.022 & 0.001 \\
\bottomrule
\end{tabular}
\caption{Agreement with human judgments on STS-test after calibration by the three maps fitted on STS-train. The ECE of raw scores is in Table~\ref{tab:raw}. $\boldsymbol{\Delta\rho}$: change in Spearman correlation caused by isotonic regression. The sigmoid and beta calibration leave $\rho$ unchanged. \textbf{Range}: difference between the highest and the lowest of the three calibrated ECE values, computed before rounding.}
\label{tab:calibrated}
\end{table}

On pairs not used for fitting each map brings the ECE of every encoder to between $0.013$ and $0.039$ (Table~\ref{tab:calibrated}). In Fig.~\ref{fig:reliability-calibrated}, the six curves lie along the diagonal and close to one another, so a calibrated score of $0.8$ is rated about $0.8$ by humans whatever the encoder. The largest gain is for \texttt{e5-base}, whose ECE falls from $0.354$ to between $0.013$ and $0.017$ depending on the map. Both the sigmoid and beta calibration leave $\rho$ exactly unchanged and the ties of isotonic regression change it by at most $0.0007$. On every encoder the three calibrated ECE values differ by at most $0.004$. Calibration brings little to \texttt{all-mpnet-base-v2}, whose raw ECE of $0.042$ is already low and stays at $0.038$ to $0.039$ after calibration. Its curve stays slightly above the diagonal.

The three maps differ only for the most similar pairs, those with a calibrated score above $0.8$. Humans give many of them the top rating of $1$, which the sigmoid cannot follow. A cosine never exceeds $1$, so the highest value the sigmoid can reach is $g(1) = 1/(1 + e^{-(a+b)})$, which is below $1$. These pairs are therefore scored too low and the sigmoid curves rise above the diagonal at the right of Fig.~\ref{fig:reliability-calibrated}. Beta calibration has no such limit, since its term $-b \ln(1-u)$ grows without bound as the cosine approaches $1$. Isotonic regression has no fixed shape and can reach $1$ too.

\section{Most of the Difference Is a Matter of Scale}
\label{sec:scale}

The second test asks whether calibration makes the encoders agree with one another, both on the value they give a pair and on the decision a threshold takes. Let $s_{e,i}$ be the cosine that encoder $e$ gives to pair $i$ of an evaluation set of $N$ pairs and $g_e$ the map fitted for encoder $e$.

\subsection{Agreement on values}

\begin{figure}[H]
\centering
\includegraphics[width=\columnwidth]{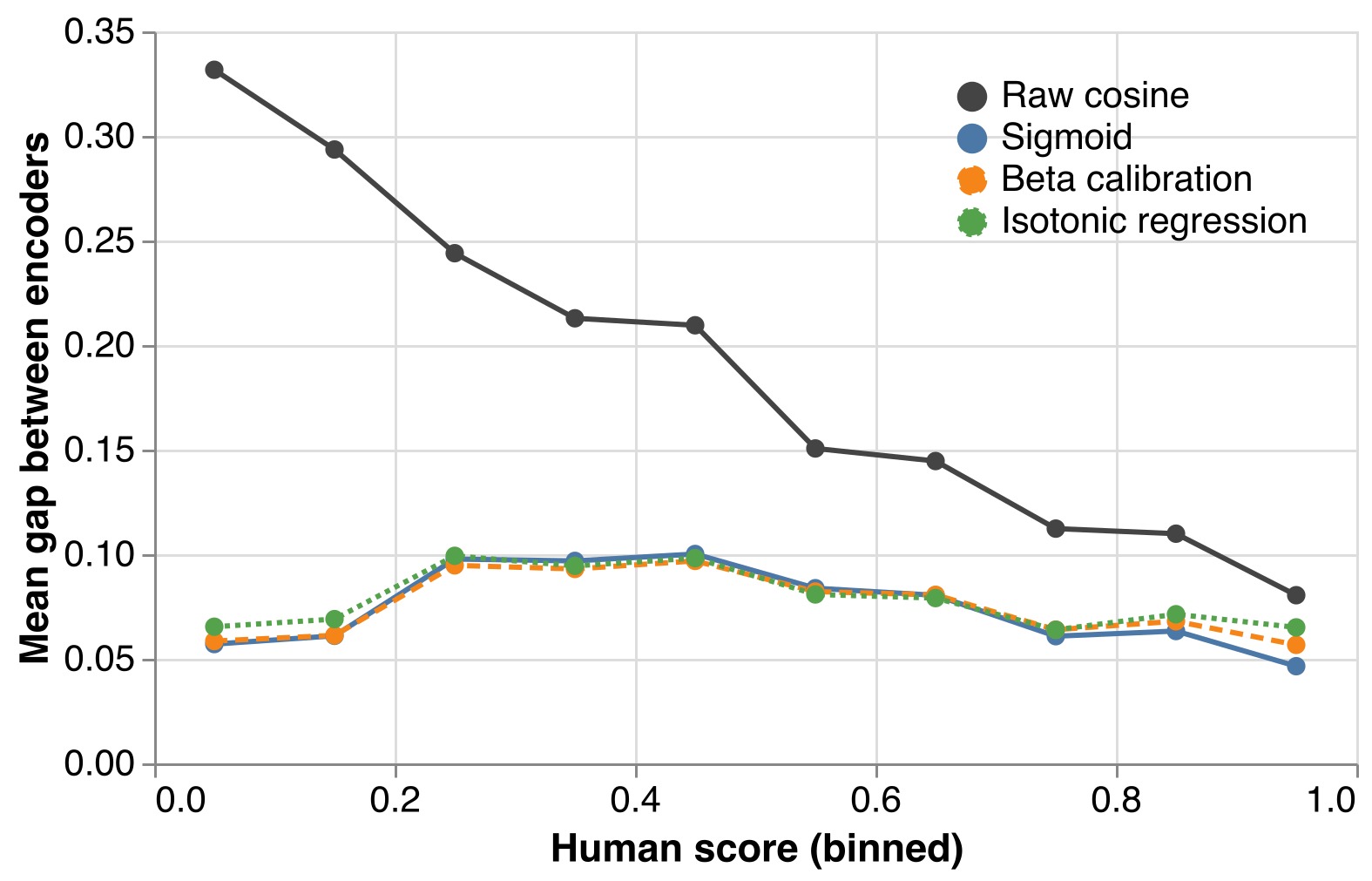}
\caption{Gap $D_k$ on STS-test, for raw scores and the three maps fitted on STS-train.}
\label{fig:gap}
\end{figure}

Agreement on values asks whether the encoders give the same score to the same pair. For pair $i$ we measure it by the mean gap between the scores of every two encoders,
\begin{equation}
\label{eq:pairgap}
d_i = \frac{1}{15} \sum_{e < e'} \lvert g_e(s_{e,i}) - g_{e'}(s_{e',i}) \rvert,
\end{equation}
where $e < e'$ means that every two distinct encoders are taken once. This gap depends on the calibration method that gives the maps $g_e$ and applies to raw cosine as well, with every $g_e$ the identity. To see how the gap depends on how similar a pair is, we split the human scale into ten bins $I_1 = [0, 0.1), I_2 = [0.1, 0.2), \dots, I_{10} = [0.9, 1]$ and let $H_k = \{ i : s^{(h)}_i \in I_k \}$ be the set of pairs whose human score falls in bin $I_k$. We average the gap in each bin,
\begin{equation}
\label{eq:gapbin}
D_k = \frac{1}{|H_k|} \sum_{i \in H_k} d_i.
\end{equation}
Averaging $d_i$ over all $N$ pairs gives a single number,
\begin{equation}
\label{eq:gap}
\Delta = \frac{1}{N} \sum_{i=1}^{N} d_i.
\end{equation}
Fig.~\ref{fig:gap} plots $D_k$ against the center of $I_k$ on STS-test and Table~\ref{tab:gap} gives $\Delta$, for raw scores and the three maps.

\begin{table}[H]
\centering
\footnotesize
\setlength{\tabcolsep}{5pt}
\begin{tabular}{lcccc}
\toprule
 & & \multicolumn{3}{c}{\textbf{Calibrated}} \\
\cmidrule(lr){3-5}
 & \textbf{Raw} & \textbf{Sigmoid} & \textbf{Beta} & \textbf{Isotonic} \\
\midrule
$\Delta$ & 0.181 & 0.073 & 0.075 & 0.078 \\
\bottomrule
\end{tabular}
\caption{Mean gap $\Delta$ on STS-test, for raw scores and the three maps fitted on STS-train.}
\label{tab:gap}
\end{table}

\begin{figure}[H]
\centering
\includegraphics[width=0.85\columnwidth]{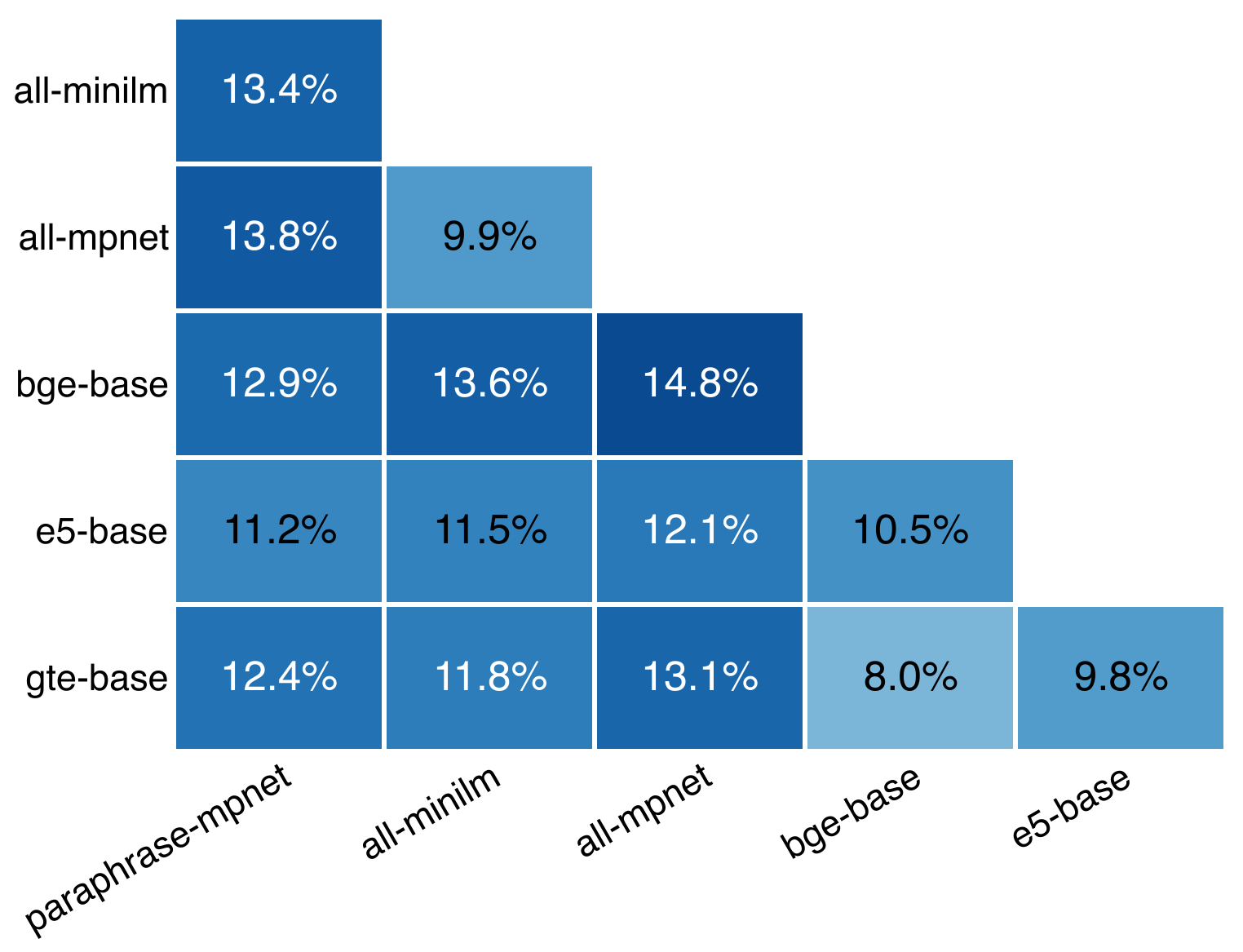}
\caption{Share of the combinations of two STS-test pairs that two encoders order in opposite ways on raw cosine, for every two encoders.}
\label{fig:order}
\end{figure}

On raw cosine the gap depends strongly on how similar a pair is. It reaches $0.332$ for pairs that humans find unrelated and falls to $0.080$ for near-paraphrases. After calibration the gap stays between $0.047$ and $0.100$ over the whole scale and the three maps give nearly the same curve. On average $\Delta$ falls from $0.181$ to $0.073$--$0.078$ depending on the map (Table~\ref{tab:gap}). The gap does not fall to zero because two encoders sometimes order two pairs in opposite ways, and no increasing map can then give them equal scores on both pairs. Fig.~\ref{fig:order} shows how often this happens, from $8.0\%$ to $14.8\%$ of the combinations of two pairs depending on the two encoders.

\subsection{Agreement on decisions}

\begin{figure}[H]
\centering
\includegraphics[width=\columnwidth]{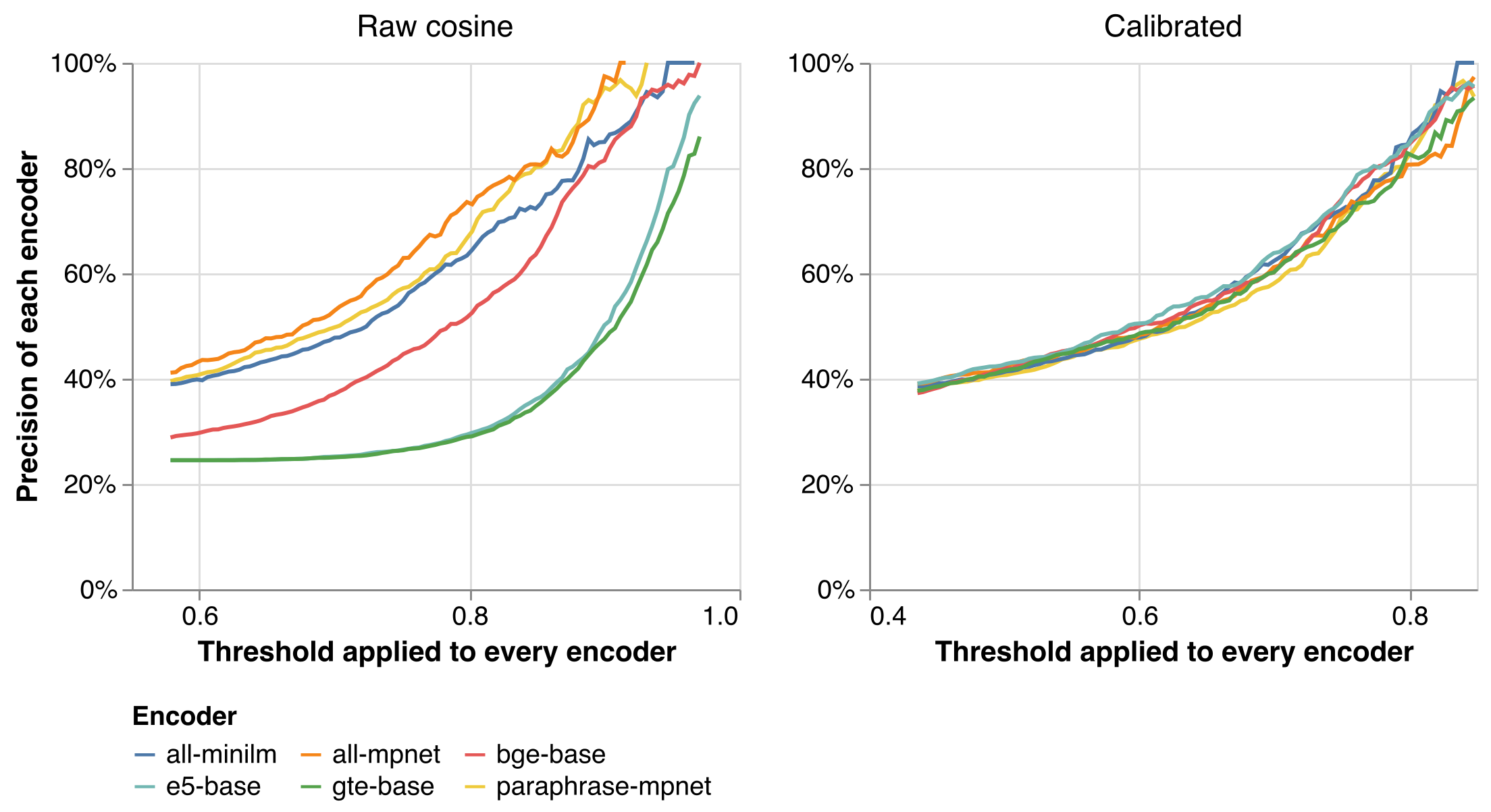}
\caption{Precision $\mathrm{prec}_e(t)$ on STS-test, on raw (left) and sigmoid-calibrated (right) scores.}
\label{fig:threshold}
\end{figure}

A similarity score is often used to decide, for instance whether two texts are duplicates or whether a cached answer can be reused. Such a decision accepts a pair when its score reaches a threshold. On raw cosine the same threshold does not mean the same thing for every encoder. A threshold of $0.8$ accepts $17.5\%$ of the test pairs with \texttt{all-mpnet-base-v2} and $84.5\%$ with \texttt{gte-base} (Table~\ref{tab:raw}), so a threshold tuned for one encoder must be tuned again whenever the encoder changes. Decisions are compared by applying one threshold $t$ to all six encoders. Encoder $e$ accepts pair $i$ when $g_e(s_{e,i}) \geq t$ and the pair is truly similar when $s^{(h)}_i \geq 0.8$. The precision of encoder $e$ is the share of accepted pairs that are truly similar,
\begin{equation}
\label{eq:precision}
\mathrm{prec}_e(t) = \frac{\big|\{ i : g_e(s_{e,i}) \geq t \wedge s^{(h)}_i \geq 0.8 \}\big|}{\big|\{ i : g_e(s_{e,i}) \geq t \}\big|}.
\end{equation}
Fig.~\ref{fig:threshold} plots it against $t$ on STS-test for raw and sigmoid-calibrated scores and Table~\ref{tab:precision} gives the precision of the three maps at a threshold of $0.8$. The other two maps are not plotted, since at this threshold the precisions of the three maps lie within $4.4$ percentage points of one another on every encoder.

\begin{table}[H]
\centering
\footnotesize
\setlength{\tabcolsep}{5pt}
\begin{tabular}{lcccc}
\toprule
 & & \multicolumn{3}{c}{\textbf{Calibrated}} \\
\cmidrule(lr){3-5}
\textbf{Encoder} & \textbf{Raw} & \textbf{Sigmoid} & \textbf{Beta} & \textbf{Isotonic} \\
\midrule
paraphrase-mpnet & 67.2\% & 83.2\% & 80.1\% & 78.9\% \\
all-MiniLM       & 63.9\% & 85.0\% & 84.9\% & 85.0\% \\
all-mpnet        & 73.4\% & 80.3\% & 78.9\% & 80.6\% \\
bge-base         & 52.3\% & 84.9\% & 83.1\% & 83.1\% \\
e5-base          & 29.6\% & 84.7\% & 83.1\% & 85.9\% \\
gte-base         & 29.0\% & 82.8\% & 78.9\% & 82.6\% \\
\bottomrule
\end{tabular}
\caption{Precision $\mathrm{prec}_e(0.8)$ on STS-test, for raw scores and the three maps fitted on STS-train.}
\label{tab:precision}
\end{table}

On raw cosine a threshold of $0.8$ reaches a precision of $29.0\%$ to $73.4\%$ depending on the encoder (Table~\ref{tab:precision}). After calibration the same threshold reaches a precision of $78.9\%$ to $85.9\%$ for every encoder, whatever the map. Fig.~\ref{fig:threshold} shows that this agreement holds at every threshold. Raw curves differ widely whatever the threshold, while calibrated curves nearly coincide.

\section{The Scale Belongs to the Reference Judgments}
\label{sec:protocol}
\begin{figure}[H]
\centering
\includegraphics[width=\columnwidth]{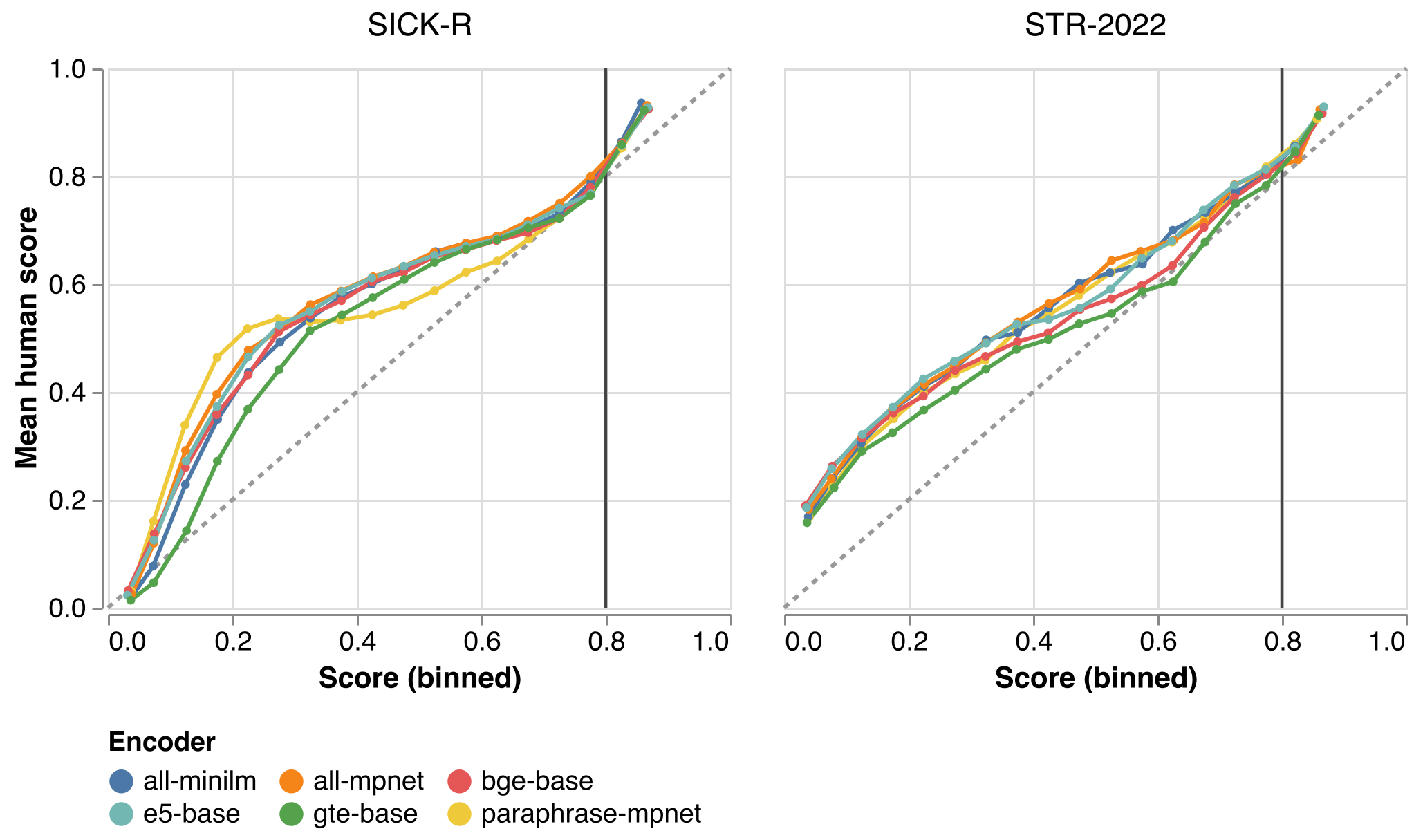}
\caption{Reliability diagram of the six encoders on SICK-R (left) and STR-2022 (right), with the sigmoid fitted on STS-train.}
\label{fig:reliability-transfer}
\end{figure}

The last test asks whether a calibration carries over to judgments collected differently. We apply the maps fitted on STS-train to two other datasets without fitting them again. On each we measure the ECE of every encoder and the mean gap $\Delta$ between encoders. SICK-R~\cite{marelli2014sick} has pairs scored for relatedness on a $1$--$5$ scale, with other sentences and other guidelines, and the release we use has $9{,}927$ of them. STR-2022~\cite{abdalla2023makes} has $5{,}500$ English pairs scored for relatedness by comparison, with annotators repeatedly picking the most and the least related of four pairs, from which a score in $[0, 1]$ is derived. We remove its $269$ pairs that share a sentence with STS-B, which leaves $5{,}231$. Scores are rescaled to $[0,1]$ from each dataset's own scale. Table~\ref{tab:transfer} gives the ECE of the three maps on both datasets and Fig.~\ref{fig:reliability-transfer} their reliability diagram with the sigmoid. The other two maps are not plotted, since the ECE of the three maps differs by at most $0.008$ on every encoder.

\begin{table}[H]
\centering
\footnotesize
\setlength{\tabcolsep}{5pt}
\begin{tabular}{lcccc}
\toprule
 & & \multicolumn{3}{c}{\textbf{Calibrated}} \\
\cmidrule(lr){3-5}
\textbf{Encoder} & \textbf{Raw} & \textbf{Sigmoid} & \textbf{Beta} & \textbf{Isotonic} \\
\midrule
\multicolumn{5}{l}{\emph{SICK-R ($9{,}927$ pairs)}} \\
paraphrase-mpnet & 0.056 & 0.086 & 0.081 & 0.083 \\
all-MiniLM       & 0.051 & 0.088 & 0.085 & 0.086 \\
all-mpnet        & 0.041 & 0.111 & 0.108 & 0.107 \\
bge-base         & 0.117 & 0.096 & 0.091 & 0.092 \\
e5-base          & 0.252 & 0.099 & 0.094 & 0.095 \\
gte-base         & 0.262 & 0.072 & 0.067 & 0.064 \\
\midrule
\multicolumn{5}{l}{\emph{STR-2022 ($5{,}231$ pairs)}} \\
paraphrase-mpnet & 0.031 & 0.125 & 0.123 & 0.124 \\
all-MiniLM       & 0.046 & 0.137 & 0.136 & 0.137 \\
all-mpnet        & 0.046 & 0.140 & 0.138 & 0.140 \\
bge-base         & 0.159 & 0.115 & 0.112 & 0.113 \\
e5-base          & 0.340 & 0.140 & 0.135 & 0.138 \\
gte-base         & 0.357 & 0.083 & 0.079 & 0.079 \\
\bottomrule
\end{tabular}
\caption{ECE of raw and calibrated scores on SICK-R and STR-2022, with the three maps fitted on STS-train.}
\label{tab:transfer}
\end{table}

On both datasets the calibrated curves of every encoder lie above the diagonal. The calibrated ECE ranges from $0.064$ to $0.111$ on SICK-R and from $0.079$ to $0.140$ on STR-2022 (Table~\ref{tab:transfer}). Calibration now underestimates human scores. The calibrated values thus do not transfer, whatever the map. Since every encoder is pushed the same way, the cause lies in the data rather than in the encoders. The annotation protocol is the most plausible one, as the same pair probably does not receive the same normalized score under the STS guidelines, the SICK relatedness guidelines and the comparative annotation of STR-2022. Our experiment cannot isolate it because the sentences and the mix of pairs change along with the protocol.

Table~\ref{tab:sickr-agreement} gives the mean gap $\Delta$ on both datasets.

\begin{table}[H]
\centering
\footnotesize
\setlength{\tabcolsep}{5pt}
\begin{tabular}{lcccc}
\toprule
 & & \multicolumn{3}{c}{\textbf{Calibrated}} \\
\cmidrule(lr){3-5}
 & \textbf{Raw} & \textbf{Sigmoid} & \textbf{Beta} & \textbf{Isotonic} \\
\midrule
SICK-R   & 0.169 & 0.074 & 0.075 & 0.078 \\
STR-2022 & 0.232 & 0.082 & 0.081 & 0.085 \\
\bottomrule
\end{tabular}
\caption{Mean gap $\Delta$ on SICK-R and STR-2022, for raw scores and the three maps fitted on STS-train.}
\label{tab:sickr-agreement}
\end{table}

The mean gap $\Delta$ falls from $0.169$ raw to $0.074$--$0.078$ calibrated on SICK-R and from $0.232$ to $0.081$--$0.085$ on STR-2022 (Table~\ref{tab:sickr-agreement}), as on STS-test. Comparability across encoders thus survives the shift, because the shift moves the calibrated scores of every encoder in the same direction. A calibration fixes what a value means for the judgments it is fitted on, and keeps encoders comparable beyond them.

\section{Discussion}
\label{sec:discussion}

Our claim is not a new calibration technique but a change in how similarity values are read. A raw cosine orders pairs but its value cannot be read in absolute terms or compared across encoders. Fitting a monotone map per encoder on a common reference fixes this without reversing the order of two pairs. The form of the map does not matter here, since three maps of very different flexibility lead to the same conclusions.

Choosing a map matters only at the top of the scale. The sigmoid stays below $1$ (Section~\ref{sec:agreement}) and no sigmoid-calibrated score of STS-test reaches $0.9$ (Fig.~\ref{fig:reliability-calibrated}), so a threshold of $0.9$ accepts no pair. Beta calibration and isotonic regression reach higher and suit thresholds above $0.85$, but isotonic regression creates ties and needs more rated pairs to place its steps.

The shift on SICK-R and STR-2022 is not uniform. In Fig.~\ref{fig:reliability-transfer} the calibrated curves depart most from the diagonal in the middle of the scale and meet it near $0.8$. One possible reading is that STS-B rates similarity while the other two rate relatedness. Two sentences can be related without saying the same thing, so relatedness exceeds similarity in the middle of the scale while near-paraphrases score high under both. Testing this would require the same sentences rated under both notions.

Semantic caches, retrieval and deduplication decide through a threshold on similarity. A calibrated threshold gives nearly the same precision whatever the encoder and should therefore carry over better when the encoder changes, although each application must check this on its own data. Such a score estimates the mean human rating of similar pairs, not the probability that a cached answer is correct.

A calibration is specific to an encoder and to the judgments it is fitted on, so an application should calibrate on judgments that match its own notion of similarity, domain and language. Calibration reduces the bias but not the noise and leaves the order of pairs unchanged up to the ties of isotonic regression.

\section{Conclusion}
\label{sec:conclusion}

A raw cosine orders pairs well, but its value has no meaning shared across encoders. Thus a cosine of $0.8$ means $0.763$ on the human scale for one encoder and $0.189$ for another. Calibrating the metric rather than the space closes most of this gap without recomputing a single embedding. An increasing map fitted per encoder on human judgments puts every encoder on the human scale, whichever of the three maps is used. One calibrated threshold then selects pairs with nearly the same precision for every encoder. The calibrated values belong to the judgments the map is fitted on and do not transfer to SICK-R and STR-2022, two datasets annotated differently, while the comparability of encoders does. Calibrated similarity can thus replace cosine similarity wherever a value is read or a threshold is set, provided it is fitted on human judgments that match the intended use.

\section*{Acknowledgments}
The author thanks the creators of the STS Benchmark, SICK and STR-2022 and of Transformers.js.

\section*{Data and Code Availability}
All datasets are publicly available, and releasing the calibration code will be considered upon publication.

\bibliographystyle{IEEEtran}
\bibliography{references}

\begin{thebibliography}{10}
\providecommand{\url}[1]{#1}
\csname url@samestyle\endcsname
\providecommand{\newblock}{\relax}
\providecommand{\bibinfo}[2]{#2}
\providecommand{\BIBentrySTDinterwordspacing}{\spaceskip=0pt\relax}
\providecommand{\BIBentryALTinterwordstretchfactor}{4}
\providecommand{\BIBentryALTinterwordspacing}{\spaceskip=\fontdimen2\font plus
\BIBentryALTinterwordstretchfactor\fontdimen3\font minus
  \fontdimen4\font\relax}
\providecommand{\BIBforeignlanguage}[2]{{%
\expandafter\ifx\csname l@#1\endcsname\relax
\typeout{** WARNING: IEEEtran.bst: No hyphenation pattern has been}%
\typeout{** loaded for the language `#1'. Using the pattern for}%
\typeout{** the default language instead.}%
\else
\language=\csname l@#1\endcsname
\fi
#2}}
\providecommand{\BIBdecl}{\relax}
\BIBdecl

\bibitem{you2026semantics}
K.~You, ``Semantics at an angle: When cosine similarity works until it
  doesn't,'' 2026, version 4.

\bibitem{schroeder2026vcache}
L.~G. Schroeder, A.~Desai, A.~Cuadron, K.~Chu, S.~Liu, M.~Zhao, S.~Krusche,
  A.~Kemper, M.~Zaharia, and J.~E. Gonzalez, ``{vCache}: Verified semantic
  prompt caching,'' in \emph{International Conference on Learning
  Representations}, 2026.

\bibitem{ethayarajh2019contextual}
K.~Ethayarajh, ``How contextual are contextualized word representations?
  {C}omparing the geometry of {BERT}, {ELMo}, and {GPT}-2 embeddings,'' in
  \emph{Proceedings of the 2019 Conference on Empirical Methods in Natural
  Language Processing and the 9th International Joint Conference on Natural
  Language Processing (EMNLP-IJCNLP)}, 2019, pp. 55--65.

\bibitem{gao2019representation}
J.~Gao, D.~He, X.~Tan, T.~Qin, L.~Wang, and T.-Y. Liu, ``Representation
  degeneration problem in training natural language generation models,'' in
  \emph{International Conference on Learning Representations}, 2019.

\bibitem{wang2020understanding}
T.~Wang and P.~Isola, ``Understanding contrastive representation learning
  through alignment and uniformity on the hypersphere,'' in \emph{Proceedings
  of the 37th International Conference on Machine Learning}, ser. Proceedings
  of Machine Learning Research, vol. 119, 2020, pp. 9929--9939.

\bibitem{su2021whitening}
J.~Su, J.~Cao, W.~Liu, and Y.~Ou, ``Whitening sentence representations for
  better semantics and faster retrieval,'' 2021.

\bibitem{platt2000probabilistic}
J.~C. Platt, ``Probabilistic outputs for support vector machines and
  comparisons to regularized likelihood methods,'' in \emph{Advances in Large
  Margin Classifiers}, A.~J. Smola, P.~Bartlett, B.~Sch{\"o}lkopf, and
  D.~Schuurmans, Eds.\hskip 1em plus 0.5em minus 0.4em\relax MIT Press, 2000.

\bibitem{zadrozny2002transforming}
B.~Zadrozny and C.~Elkan, ``Transforming classifier scores into accurate
  multiclass probability estimates,'' in \emph{Proceedings of the Eighth ACM
  SIGKDD International Conference on Knowledge Discovery and Data Mining},
  2002, pp. 694--699.

\bibitem{niculescumizil2005predicting}
A.~Niculescu-Mizil and R.~Caruana, ``Predicting good probabilities with
  supervised learning,'' in \emph{Proceedings of the 22nd International
  Conference on Machine Learning}, 2005, pp. 625--632.

\bibitem{kull2017beta}
M.~Kull, T.~Silva~Filho, and P.~Flach, ``Beta calibration: A well-founded and
  easily implemented improvement on logistic calibration for binary
  classifiers,'' in \emph{Proceedings of the 20th International Conference on
  Artificial Intelligence and Statistics}, ser. Proceedings of Machine Learning
  Research, vol.~54, 2017, pp. 623--631.

\bibitem{guo2017calibration}
C.~Guo, G.~Pleiss, Y.~Sun, and K.~Q. Weinberger, ``On calibration of modern
  neural networks,'' in \emph{Proceedings of the 34th International Conference
  on Machine Learning}, ser. Proceedings of Machine Learning Research, vol.~70,
  2017, pp. 1321--1330.

\bibitem{desai2020calibration}
S.~Desai and G.~Durrett, ``Calibration of pre-trained transformers,'' in
  \emph{Proceedings of the 2020 Conference on Empirical Methods in Natural
  Language Processing (EMNLP)}, 2020, pp. 295--302.

\bibitem{cer2017semeval}
D.~Cer, M.~Diab, E.~Agirre, I.~Lopez-Gazpio, and L.~Specia, ``{SemEval}-2017
  task 1: Semantic textual similarity multilingual and crosslingual focused
  evaluation,'' in \emph{Proceedings of the 11th International Workshop on
  Semantic Evaluation (SemEval-2017)}, 2017, pp. 1--14.

\bibitem{muennighoff2023mteb}
N.~Muennighoff, N.~Tazi, L.~Magne, and N.~Reimers, ``{MTEB}: Massive text
  embedding benchmark,'' in \emph{Proceedings of the 17th Conference of the
  European Chapter of the Association for Computational Linguistics}, 2023, pp.
  2014--2037.

\bibitem{vershynin2018high}
R.~Vershynin, \emph{High-Dimensional Probability: An Introduction with
  Applications in Data Science}.\hskip 1em plus 0.5em minus 0.4em\relax
  Cambridge University Press, 2018.

\bibitem{wang2022text}
L.~Wang, N.~Yang, X.~Huang, B.~Jiao, L.~Yang, D.~Jiang, R.~Majumder, and
  F.~Wei, ``Text embeddings by weakly-supervised contrastive pre-training,''
  2022.

\bibitem{levenberg1944method}
K.~Levenberg, ``A method for the solution of certain non-linear problems in
  least squares,'' \emph{Quarterly of Applied Mathematics}, vol.~2, no.~2, pp.
  164--168, 1944.

\bibitem{marquardt1963algorithm}
D.~W. Marquardt, ``An algorithm for least-squares estimation of nonlinear
  parameters,'' \emph{Journal of the Society for Industrial and Applied
  Mathematics}, vol.~11, no.~2, pp. 431--441, 1963.

\bibitem{ayer1955empirical}
M.~Ayer, H.~D. Brunk, G.~M. Ewing, W.~T. Reid, and E.~Silverman, ``An empirical
  distribution function for sampling with incomplete information,'' \emph{The
  Annals of Mathematical Statistics}, vol.~26, no.~4, pp. 641--647, 1955.

\bibitem{marelli2014sick}
M.~Marelli, S.~Menini, M.~Baroni, L.~Bentivogli, R.~Bernardi, and
  R.~Zamparelli, ``A {SICK} cure for the evaluation of compositional
  distributional semantic models,'' in \emph{Proceedings of the Ninth
  International Conference on Language Resources and Evaluation (LREC'14)},
  2014, pp. 216--223.

\bibitem{abdalla2023makes}
M.~Abdalla, K.~Vishnubhotla, and S.~Mohammad, ``What makes sentences
  semantically related? a textual relatedness dataset and empirical study,'' in
  \emph{Proceedings of the 17th Conference of the European Chapter of the
  Association for Computational Linguistics}, 2023, pp. 782--796.

\end{thebibliography}

\end{document}